# Extraction of Cartographic Objects on High Resolution Satellite Images for Object Model Generation

Guray Erus, Nicolas Loménie
*SIP-CRIP5 Laboratory, Université de Paris 5*
{egur,lomenie}@math-info.univ-paris5.fr

**Abstract**

*The aim of this study is to detect man-made cartographic objects in high-resolution satellite images. New generation satellites offer a sub-metric spatial resolution, in which it is possible (and necessary) to develop methods at object level rather than at pixel level, and to exploit structural features of objects. With this aim, a method to generate structural object models from manually segmented images has been developed. To generate the model from non-segmented images, extraction of the objects from the sample images is required. A hybrid method of extraction (both in terms of input sources and segmentation algorithms) is proposed: A region based segmentation is applied on a 10 meter resolution multi-spectral image. The result is used as marker in a "marker-controlled watershed method using edges" on a 2.5 meter resolution panchromatic image. Very promising results have been obtained even on images where the limits of the target objects are not apparent.*

## 1. Introduction

With the spread of very high-resolution satellite images, more sophisticated image processing systems are required for the automatic extraction of information. In the frame of a research project of CNES[1] to develop tools and algorithms to exploit images acquired by the new generation Pleiades satellites, a database of "cartographic object images" has been prepared. This database will be used to detect the target objects in a global scene, on a very large satellite image. For this task, low-level pixel based classification methods are not very successful, mainly because they disregard the structural features of the target objects. Man-made geographical objects have well-defined (but quite variable), mostly geometrical structures. In low-resolution, the objects disappear and become part of the texture. However, in high resolution, the shape features and the spatial relations between the objects are perceivable and exploitable.

As a first module, a detection method that uses very basic structural and spatial features of the objects has been developed [1]. This method generates an object model from the given examples. As input, the system uses sample images of objects manually segmented by an expert. The images are transformed to Attributed Relational Graphs (ARGs). A model representing the common features of the objects is constructed applying graph-matching algorithms [2].

In order to generate the model from the initial satellite images, the preliminary task is the extraction of the objects from the sample images (as done by the manual segmentation). The satellite images of the same object in two different spectral bands and resolutions are provided (Fig 1): A panchromatic image ($I_p$) with resolution of 2.5 meters, and a multispectral ($I_m$) image with resolution of 10 meters.

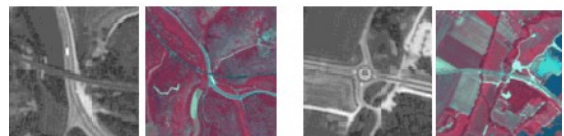

*Fig 1. Ip and Im images of a bridge and a round-about*

This paper describes a hybrid segmentation approach that uses the two images of different resolutions, and that combines region-based and edge-based methods by a "marker-controlled watershed method using edges". The originality of this method lies in the fusion of complementary information from different sources and methods to obtain a good segmentation.

The ultimate objective is to combine the two modules and to integrate them in an interactive object detection software based on query by example. The object model learned from the examples will be used as a semantic filter (together with a radiometric filter) in the detection of the candidate regions.

---
[1] French National Space Agency

## 2. Model Generation

The segmented images are first split into primitives either using simple geometric shapes (rectangles and circles) or using the skeleton of the objects. They are then transformed into ARGs. The features of primitives are stored in the vertices of the graph and the features of connections in the edges. The vertex attributes are the type of the object and the edge attributes are the type and direction of the connection.

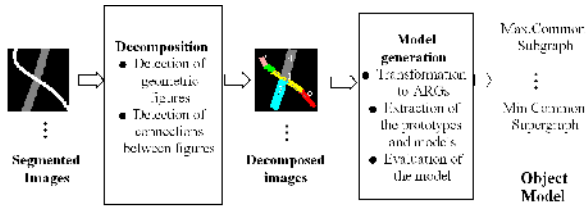

*Fig 2. Model generation*

From the ARGs belonging to a class of object, the prototypes, the most common representations of each class, are detected using an exact matching algorithm. The model is obtained by finding the Maximal Common Subgraph (*MaxCSg*) and the Minimal Common Supergraph (*MinCSg*) from the prototypes of an object. The main steps of the algorithm are given in figure 2.

To evaluate the quality of a model, the metric proposed by Bunke [3] to calculate the edit distance between two graphs is adapted to the distance between an ARG and the model.

Figure 3 presents the models generated for the bridge and round-about objects. The models are simple and quite similar to manually generated models. They represent the geometrical and spatial characteristics of the target objects.

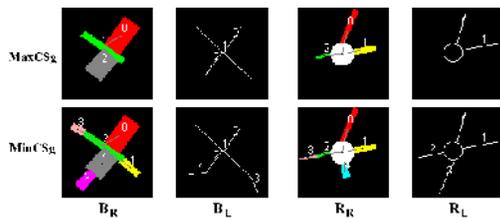

*Fig 3. Bridge and round-about models: using rectangles and circles, using line segments and circles*

The manually segmented images are not always available. In order to use the method on a real world system, the target objects should be first extracted from the background. Consequently, our work is concentrated on the extraction of target objects from the satellite images.

## 3. Extraction of Target Objects

Extraction of objects on $I_p$ images using naive methods does not give a satisfactory result due to the imprecision of the boundaries separating cartographic objects from the background. On the other hand, while the objects can be well differentiated from the background on $I_m$ images, their shape is too coarse. The use of both images makes it possible to overrun these limitations.

The steps of the method are given in the following sections.

### 3.1. Segmentation of the $I_m$ Image.

The first step is to clip the central part of the $I_m$ image and to magnify it by a factor of 4, so that its resolution becomes the same as the resolution of the $I_s$ image. A linear interpolation is applied to magnify the image.

The $I_m$ image consists of three channels $I_1$, $I_2$, $I_3$ with spectral ranges 0.50-0.59 μm, 0.61-0.69 μm and 0.78-0.89 μm. The target objects are consisting of pieces of possibly different types of roads in a specific spatial arrangement. We generated the linear combination of the three channels, that discriminates better the roads from other objects, fixing the coefficients empirically. A deeper analysis in parameter space combined with prior knowledge on reflectance values of spectral bands may result in a much better discrimination. The resulting image $I_f$ is obtained by:

$$I_f = (I_1 + I_2) * 0.3 - I_3$$

On $I_f$ a hysteresis threshold is applied to extract the region $R_M$ containing the target object. The threshold value is calculated by finding the most frequent gray value in a 5x5 central window on all $I_m$ images. The method is very successful in detecting the true positives (Fig. 4). There are some false positives due to the incorrect threshold value for some of the images, but these are acceptable considering that the result will be refined in the next steps.

### 3.2. Matching the Mask.

It is necessary to search for the exact location of the mask on the $I_s$ image. There exist many sophisticated image registration techniques [4]. In our case, we don't deal with the transformations and deformations that the image has been subject to, and the difference between centers is limited to at most 10 pixels in horizontal and in vertical in both directions. A matching method moving the center of the mask in a 20x20 window located on the center of the $I_s$ image would be sufficient to visit all "candidate" points.

For each candidate point $p_i$ a matching score should be calculated. The mask, in the correct place, should contain the whole target object. The most pertinent information that indicates the location of the object is given by the edges. A correctly placed mask that contains an object should also contain its edges. Besides that, the edges in the exterior part of an object would remain out of the mask if the mask is not well placed. Consequently, the score is calculated using the number of edge points contained in a region.

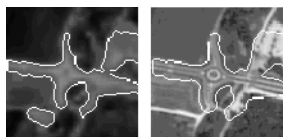

*Fig. 4. Mask on Im image and on Ip image after matching*

Sub-pixel precision edges are obtained using Canny edge detection algorithm [5]. In order to eliminate noisy short edge segments, the edges are smoothed and edges with closer extremities are merged. Finally short edges are eliminated. Before calculating the score, a dilation is applied on the mask so that it would be slightly bigger than the object. In case of equal score, the location that gives the smallest variance score for the masked region is selected. The mask center is shifted to the point with the highest score (Fig. 4).

### 3.3. Extraction of the object.

The region covered by the mask is an approximation of the object shape obtained from a lower resolution image. The mask should be refined in order to extract the object.

Region-based methods fail to locate the object boundaries well. It is also difficult to select the regions belonging to the object. Edge-based methods give us a good localization of the separation lines between the object and the background. However, the extracted edges are not continuous, and consequently they don't give us closed regions.

It is a very common approach to integrate region growing and edge detection methods, as Pavlidis et al. [6] that uses the edge intensities to eliminate the boundaries in an over- segmented image, or Le Moigne et al. [7] that uses the edge features to determine the stopping criterion of the region growing algorithm.

In his multi-scale segmentation algorithm, [8] proposed to increase the intensities of the straight edge segments obtained from a different source, on the gradient image before applying the watershed. In that way these segments are preserved in the final watersheds.

The method proposed in this paper is a marker-controlled watershed segmentation that uses the edge information as in [8].

Watershed method produce an important over-segmentation. A solution is to apply a fusion algorithm. Since we look for a unique object, a very appropriate alternative is to use markers to limit the number of resultant regions. The main difficulty is to detect the markers. Usually, this is done through user interaction. In our case, a mask covering the object is already obtained. It can be assumed that the skeleton of the mask should match with the skeleton of the object, consequently it is a good marker for the object. Similarly the external boundaries of the mask can be used to mark the background.

The steps of the final algorithm are as follows:
1. Detect:
   $S_m$ = skeleton of the mask M.
   $B_m$ = the external boundary of M after dilation.
   $Ig$ = gradient of Is
   $Es$ = edges of Is (after previously mentioned post processing)
2. Set Ig(Es) to max(Ig) (Pixels belonging to the edges will have the highest gradient values)
3. Modify Ig so that it has minima only at $S_m$ and $B_m$.
4. Apply the watershed.

The algorithm gives a unique region as the object around the skeleton bounded by the detected edges.

### 4.3. Experimental Results

We tested our system on 20 bridge and 20 round-about objects.

| Step | Object | Results | | |
|---|---|---|---|---|
| | | Correct | Acceptable | Incorrect |
| Segm. | Bridge | 15 | 3 | 2 |
| | Round-about | 10 | 5 | 5 |
| Match. | Bridge | 14 | 1 | 5 |
| | Round-about | 10 | 3 | 7 |
| Extract. | Bridge | 8 | 6 | 6 |
| | Round-about | 6 | 5 | 9 |

*Table 1. The cumulative results*

We evaluated the results by manual inspection in the end of each step and by comparing them with images segmented by an expert at the end of the whole process. Table 1 gives a brief overview of the results at the end of each step.

The results of the first step are very successful for the bridges. In almost all of the images the object is partially or totally detected. For the round-abouts there are some bad segmentations mainly due to two reasons: the small size of the object in some images makes the segmentation very difficult in the low resolution image and buildings very near to the round-abouts are confused with the roads. In the second step, the rate of exact match is very high. The incorrect matches occur mainly when the mask size is large. After eliminating the errors from the previous steps, the results of the extraction phase are very successful.

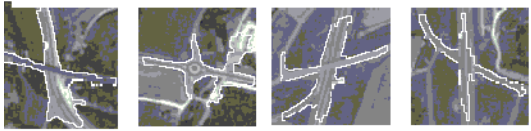

*Figure 5. The extracted objects*

A qualitative analysis seems us more appropriate for the final step. The segmentation follows exactly the edge lines and connects them according to the maximum values of the gradient values in disconnected regions. This behavior is conform to what we aimed using a hybrid approach.. However, the result is highly dependent on the skeleton obtained from the mask, and when there are multiple edges it may follow the incorrect edges (Fig. 5). Defining a weight on the length and the position of the edge line may help to solve this problem.

## 4. Conclusions

This study is a first approach to a difficult problem, and we proposed a system that can be refined with several improvements. The extraction module gives promising results. However, the results should be improved in order to apply the model generation directly on them.

Detection of circles and line segments may help correcting the irregularities of the final segmentation. For the round-about images, the early detection of the central circle using Hough transformation may be used to improve the results of the first two steps. Another alternative is to implement deformable models.

We can propose several refinements and ameliorations for the model generation. Considering the numerical attributes, using a fuzzy modelisation of the symbolic concepts and integrating methods issued from qualitative spatial reasoning as in [9] may enhance the results of our system.

Another important perspective is to iterate the segmentation process using the generated model so that the high-level knowledge would be used in low-level processing.

## 5. Acknowledgements

We would like to thank French Space Agency - CNES, Toulouse, and especially Gilbert Pauc and Jordi Inglada for their support. We would also like to thank Dr. Z. Hamrouni and Cultural Service of the French Embassy in Ankara.